\pdfoutput=1

\documentclass[11pt]{article}
\usepackage[]{emnlp2022}
\usepackage{emnlp2022}
\usepackage{times}
\usepackage{latexsym}
\usepackage{amsmath}
\usepackage{titlesec}
\usepackage{xcolor}         
\usepackage{graphicx,amssymb,amsmath,bm}
\usepackage[export]{adjustbox}
\usepackage{multirow}
\usepackage{booktabs}
\usepackage{multirow}

\usepackage[T1]{fontenc}

\usepackage[utf8]{inputenc}

\usepackage{microtype}

\usepackage{todonotes}
\usepackage{enumitem}
\usepackage{url}

\title{\textsc{EnDex}: Evaluation of Dialogue Engagingness at Scale}

\author{Guangxuan Xu$^{1}$ \quad Ruibo Liu$^{2}$ \\ \quad \textbf{Fabrice Harel-Canada}$^{1}$ \quad \textbf{Nischal Reddy Chandra}$^{1}$ \quad \textbf{Nanyun Peng}$^{1}$
\\
  $^{1}$University of California, Los Angeles  \qquad
  $^{2}$Dartmouth College \\
   {\tt \{gxu21, violetpeng\} @cs.ucla.edu} 
}

\begin{document}
\maketitle
\begin{abstract}

We propose \textsc{EnDex}, the first human-reaction based model to evaluate dialogue engagingness. \textsc{EnDex} is trained on 80k Reddit-based Engagement Dataset (RED) curated using a novel distant-supervision framework. 
Engagingness is a key measure that captures high-level quality of AI dialogue systems and closely reflects actual user experience. However, data shortage, plus the abstract and extensive definition of engagingness makes it challenging to develop an automatic metric. Our work departs from mainstream approaches that use synthetic negative examples to train binary classifiers, and instead, proposes a solution using distant-supervision from human-reaction feedback. 
To support the soundness of our \textsc{EnDex} metric, we offer a theoretical foundation for engagement, an extensive ablation study, and empirical evidence of high correlation on five engagingness related datasets.\small\footnote{Off-the-shelf \textsc{EnDex} model and the RED dataset is available at \url{https://github.com/gxxu-ml/EnDex}.}

\end{abstract}

\section{Introduction}
Many modern generative language models are trained to maximize a likelihood objective, but this paradigm tends to assign high probability to generic responses~\citep{li2015diversity}, such as ``I don't know.''. Prior research has established that people prefer to converse with interesting, creative, and informative agents~\citep{See2019WhatMA}, all concepts broadly related to the notion of \emph{engagingness}.  Furthermore, engagingness is recognized as a key evaluation metric for the quality of dialogue systems~\citep{Zhang2018PersonalizingDA,Ghazarian2020PredictiveEA}. For example, FAIR's ParlAI~\citep{Miller2017ParlAIAD} incorporated Engagingness as the default testing metric in the Blenderbot system~\citep{Roller2021RecipesFB}; dialogue data challenges, like ConvAI2~\citep{Dinan2019TheSC}, Amazon Alexa Prize\footnote{https://www.amazon.science/alexa-prize}, and ensemble metrics like FED~\citep{Mehri2020UnsupervisedEO}, all measure engagingness to benchmark dialogue quality. 
 



\begin{figure}[t]
    \centering
    \includegraphics[width=0.45\textwidth]{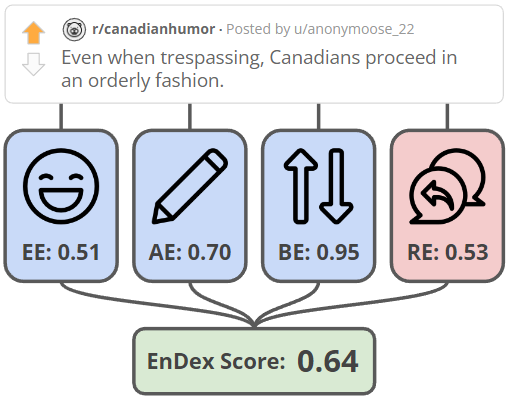}
    \caption{Example of an online post with scores for emotional engagement (EE), attentional engagement (AE), and behavioral engagement (BE) in blue to represent the 3 dimensions of \emph{human engagement}; reply engagement (RE) in red; and the aggregated \textsc{EnDex} score in green. We apply z-score to EnDex Score and pick a hyper-parameter threshold to cluster posts into positive and negative samples.}
    \vspace{-0.15in}
    \label{fig:4_dimensions}
\end{figure}


However, the current evaluation of engagingness still primarily relies on expensive human annotation rather than off-the-shelf automatic tools, due to several theoretical and technical challenges: firstly, unlike more well-characterized properties such as fluency, the definition of engagingness is significantly more abstract and multi-dimensional~\citep{See2019WhatMA}, requiring well-tuned quality metrics for each sub-dimension to aggregate a final score. Secondly, what qualifies as engaging is open-ended and many different answers may embody the concept~\citep{Ghazarian2020PredictiveEA}. Therefore, reference-based metrics requiring unique ground truth, such as BLEURT~\citep{Sellam2020BLEURTLR} and BERTScore~\citep{zhang2019bertscore}, cannot apply. Thirdly, there's an acute shortage of large-scale, high-quality data annotated for engagingness. 


\citet{Ghazarian2020PredictiveEA} jump-started efforts to automatically measure dialogue engagement, where they fine-tuned a BERT-based model \cite{bert} on the ConvAI2 and DialyDialog datasets~\citep{li-etal-2017-dailydialog} to predict an engagingness score. However, finetuning on small size supervised data could easily lead to overfitting and generalization problems. Another high performing metric on engagingness USL-H~\citep{Phy2020DeconstructTR} assumes a positive set and generates synthetic negative samples to train model. However, credible positive samples are not always available, and synthetic negative samples may not be challenging enough to further advance classifier performance.

In light of the above challenges, we propose \textsc{EnDex}, a novel metric trained with distantly supervised data to predict turn-level dialogue engagingness (Figure~\ref{fig:4_dimensions}). \textsc{EnDex} requires neither human annotations nor direct disentanglement of engagingness. Instead, we leverage observed \emph{user reactions} to posts as distant signals to model engagingness, which marks a departure from mainstream approach to train on synthetic negative samples~\citep{Lan2020PONEAN, Ghazarian2022DEAMDC, Tao2018RUBERAU,Sato2020EvaluatingDG}. \textsc{EnDex} trains on real conversations sourced from Reddit, that are automatically annotated as positive and negative examples with our framework. The novoel dataset is named \textsc{RED} (Reddit Engagement Dataset) with over 80k labelled samples. \textsc{EnDex} framework derives its theoretical underpinning from relevant HCI works, and has shown superior performance on five benchmark datasets.

\section{EnDex Metric}
Engagingness is not only a linguistic concept useful for dialogue systems, but also manifests itself in multi-modalities and is extensively leveraged to benchmark gaming and online learning experiences~\citep{inproceedings_Silpasuwanchai,Chen2005ModelingAM,articlemcmahan,SchoenauFog2011ThePE}. Our work is inspired by HCI study of \textit{Human Engagement}~\cite{inproceedings}, which decomposes engagingness into three major dimensions including \textit{attentional engagement} (e.g., clicks and scrolls), \textit{behavior engagement} (e.g., facial expressions), and \textit{emotional engagement} (e.g., heart rate).


\textsc{EnDex} metric follows the same intuition: we can infer engagingness of a text by analyzing human reactions to it, for which there is abundant data in social media. \textsc{EnDex} metric learns from our distant-supervised \textsc{RED} dataset, which measures dialogue engagement along four dimensions as shown in Figure~\ref{fig:4_dimensions}; three-dimensions correspond to the original Human Engagement definition, and one distinct Reply Engagement dimension for the dialogue specific task. 

\subsection{Reddit Engagement Dataset (\textsc{RED})}

We curate the Reddit Engagement Dataset (\textsc{RED}), a distant-supervision set, with 80k single-turn conversations. We source \textsc{RED} from Reddit, sampling from 43 popular subreddits, and processed a total of 5 million posts, filtering out data that was either non-conversational, toxic, or posts not possible to ascertain popularity; the resulting data distribution of \textsc{RED} is shown in Table~\ref{tab:data_dist}. The following sections will explain the procedure to automatically annotate \textsc{EnDex} scores and cluster samples into positive and negative sets.

We also curated a RED testset with 150 human annotated samples obtained from a different split from \textsc{RED}. The inter-annotator agreement is 0.34 Fleiss-Kappa, indicating fair agreement, which reflects the challenge of determining engagingness.




\begin{table}[t]
\centering
\resizebox{0.4\textwidth}{!}{%
\begin{tabular}{@{}lcc@{}}
\toprule
              & Engaging    & Non-engaging \\ \midrule
\# of samples & 40,162      & 40,162       \\ \cmidrule(r){1-1}
Emotional            & .605 $\pm$ .273 & .152 $\pm$ .120  \\
Attentional            & .759 $\pm$ .127 & .203 $\pm$ .100  \\
Behavioral            & .659 $\pm$ .274 & .318 $\pm$ .285  \\
Reply            & .718 $\pm$ .154 & .354 $\pm$ .980   \\ \cmidrule(r){1-1}
\textsc{EnDex}   & .709 $\pm$ .048 & .259 $\pm$ .033  \\ \bottomrule
\end{tabular}%
}
\caption{\textsc{RED} dataset has two classes, engaging and non-engaging, clustered by applying z-score on \textsc{EnDex} score. This table shows the mean and standard deviation of sub-dimension scores for both classes; the last row displays the distribution of the overall \textsc{EnDex} score.}
\label{tab:data_dist}
\end{table}




\subsection{Distantly-Supervised Engagingness Scores} \label{sec:dims}

We use distant-supervision to provide samples in \textsc{RED} an \textsc{EnDex} Score, which is the aggregate of 4 engaging dimensions. Section \ref{sec:dims} discusses the intuition for each engagingness dimension; section \ref{meth:formula} explains how to adjust raw score by thread popularity; section \ref{meth:submod} lays out the formula to normalize and aggregate sub-dimensions into the overall engagingness score; section \ref{meth:zscore} explains sampling with z-score to convert the task into binary classification. 

\begin{itemize}[leftmargin=*]

\itemsep-.3em

\item \textbf{Emotional Engagement (EE):} Emotional connection is a key sign of human engagement~\citep{SavinBaden2014StudentsEO}; and we model EE using a multi-class emotional classifier~\citep{demszky-etal-2020-goemotions} on post replies. If post receives positive and emotional replies, it's engaging; negative or neutral replies indicates non-engaging. 
\item \textbf{Attentional Engagement (AE):} More user time spent indicates higher engagement~\citep{Attfield2011TowardsAS}. We model AE of a post by examining whether it has editted replies, and the information specificity in the replies.

\item \textbf{Behavioral Engagement (BE):} Human behavioral features closely correlate with their engagement state~\citep{Attfield2011TowardsAS}, and we model BE by examining Reddit post scores, adjusted by popularity.


\item \textbf{Reply Engagement (RE):} Following definition from ~\citep{Ghazarian2020PredictiveEA}, if a certain post is very likely to be continued by following threads, it is considered engaging; reply\_counts are also popularity adjusted. 

\end{itemize}



\subsection{Adjustment for Popularity}
\label{meth:formula}
Raw score for Behavior Engagement(upvotes) and Reply Engagement(reply counts) are heavily influenced by the popularity of the thread in which the post appears. A non-engaging post may receive high user interaction because it simply receives a lot of exposure; on the flip side, a very engaging most may receive zero user interaction simply because it is rarely seen. To mitigate the imbalanced exposure problem, we calculate a popularity value for each thread, and adjust posts scores by the popularity value of the thread it resides.

\textbf{Popularity Value(PV)} The PV of a post is given by the amount of exposure its parent post attracts. Let the target post be $\theta$ and its parent $\sigma$, $R_{eply}$ obtains the reply counts of a post, and $U_{pvote}$ obtains the upvotes of a post. The PV is defined in equation (\ref{eq:pv}), where coefficient 2 is adopted to give equal weight for reply and upvotes; popularity value adjusted RE score is given by PVRE in equation (\ref{eq:pvre}), where $M_{pv}$ and $M_{re}$ are the median of popularity value and reply counts in the entire dataset. Only popularity adjusted scores are used for calculating \textsc{EnDex} score.

 

 \begin{equation}
    \small
    \label{eq:pv}
     \textrm{PV}(\theta) = 2* R_{eply}(\sigma) + U_{pvote}(\sigma)
 \end{equation}

 \begin{equation}
 \small
    \label{eq:pvre}
     \textrm{PVRE}(re) = re + \frac{M_{pv}}{M_{\textrm{re}}}  *  \frac{re}{\textrm{PV(re)}} * re
 \end{equation}

\subsection{Monotone Submodular Normalization}
\label{meth:submod}
The final \textsc{EnDex} score is essentially a weighted sum of the 4 respective sub-dimension scores; an importance nuance is the usage of submodular normalization (shown in Eq. \ref{eqn:eqn}) for 3 dimensions to bring raw scores to the scale of 0-1. We observe that unit increase in raw score lead to diminishing positive effect on engagingness. For example, a sentence with 100 replies should be more engaging than one with 1 replies, but not 99 times more; thus, we normalize engagingness score with a monotone submodular function $f(x)=\frac{x}{x+\alpha}$. 




\begin{equation}
\small
    \label{eqn:eqn}
     N(x) = \Bigg(\displaystyle\sum_{n=1} ^{3} w_i * \frac{x_i}{x_i+ \alpha_i} \Bigg) +  w_{\textrm{EE}} * x_{\textrm{EE}},
\end{equation}

\noindent $N$ is the normalized score for sample $x$, $x_i$ is $x$'s raw score on dimension $i$, where $i\in \{RE,BE,AE\}$; $\alpha_i$ is the median of $i$-th dimension; $w_i$ is the weight for $i$ dimension; $w_{\textrm{EE}}$ is the weight for EE dimension. The weight can be tuned for your own usage of \textsc{RED}; \footnote{For \textsc{EnDex}, $\alpha$ for three dimensions RE, BE, AE are 1, 2, 18, respectively; we also applied weights of 3, 3, 2, 1 for RE, AE, EE, and BE}.

\subsection{Clustering with z-score}
\label{meth:zscore}
Essentially, engagingness prediction is a classification task, and we want to prepare dataset for binary classification. We use $z$-score on the \textsc{EnDex} Score to easily sample and cluster the data according to standard deviation from mean. A confidence threshold $\kappa$(ours is 1) needs to be picked, which means that we regard samples that fall between $\kappa$ standard deviation from mean as uncertain, and are thus discarded. And we cluster positive and negative samples using the following equation (\ref{eq:zscore}).\\
\begin{equation}
\label{eq:zscore}
\small Polarity(x) = \begin{cases}
1 & \text{if } z\_score(x)>\kappa \\
0 & \text{if } z\_score(x)<-\kappa 
\end{cases}
\end{equation}

The EnDex metric is then trained as a binary classification task by finetuning a RoBERTa-large model~\citep{Liu2019RoBERTaAR} on turn-level \textsc{RED} data.



\section{Experiments}



\begin{table*}[]
\centering
\small
\renewcommand{\tabcolsep}{1.8mm}
\begin{tabular}{@{}lllllllllll@{}}
\toprule
\multirow{2}{*}{\textbf{Method}} &
  \multicolumn{2}{c}{\textsc{Better}} &
  \multicolumn{2}{c}{\textsc{PredEng-600}} &
  \multicolumn{2}{c}{\textsc{Fed-eng}} &
  \multicolumn{2}{c}{\textsc{Red-Test}} &
  \multicolumn{2}{c}{\textsc{Grade}} \\ \cmidrule(l){2-11} 
 &
  \multicolumn{1}{c}{P} &
  \multicolumn{1}{c}{S} &
  \multicolumn{1}{c}{P} &
  \multicolumn{1}{c}{S} &
  \multicolumn{1}{c}{P} &
  \multicolumn{1}{c}{S} &
  \multicolumn{1}{c}{P} &
  \multicolumn{1}{c}{S} &
  \multicolumn{1}{c}{P} &
  \multicolumn{1}{c}{S} \\ \midrule
Random (\textit{ref.})             & 0.025              & 0.025              & -0.012            & -0.013            & 0.080             & 0.081             & -0.053              & -0.053              & 0.053             & 0.045              \\
Question                           & 0.167              & 0.167              & 0.073             & 0.074             & \textbf{0.320}    & \underline{0.320} & 0.194               & 0.194               & 0.009             & 0.008              \\
Specificity                        & 0.357              & 0.357              & 0.076             & 0.102             & 0.254             & 0.254             & 0.122               & 0.122               & -0.090            & -0.090             \\
\cmidrule(r){1-1}    
USL-H                              & 0.356              & 0.343              & \textbf{0.688}    & \textbf{0.699}    & 0.267             & 0.277             & 0.121               & 0.125               & 0.234             & 0.243              \\      
\textsc{Pred\_En }                 & 0.234              & 0.310              & -0.137            & -0.134            & 0.250             & \textbf{0.340}    & 0.044               & 0.178               & -0.090            & -0.060             \\
\textsc{Pred\_En } (FT+DD)         & 0.338              & 0.368              & 0.390             & 0.450             & 0.253             & 0.195             & 0.237               & 0.258               & 0.194             & 0.173              \\ 
\cmidrule(r){1-1}    
\textbf{Ours:} \textsc{EnDex}      & 0.414*             & 0.397*             & 0.397             & 0.348             & 0.235*            & 0.225*            & 0.381**             & 0.375**             & 0.266             & 0.248              \\
\textbf{Ours:} \textsc{EnDex+NS}   & \underline{0.478}* & \underline{0.455}* & 0.597**           & 0.577**           & 0.229*            & 0.214*            & \underline{0.389}** & \underline{0.378}** & \underline{0.308} & \underline{0.282}* \\ 
\textbf{Ours:} \textsc{EnDex-Best} & \textbf{0.521}     & \textbf{0.511}     & \underline{0.620} & \underline{0.629} & \underline{0.286} & 0.253             & \textbf{0.414}      & \textbf{0.405}      & \textbf{0.406}    & \textbf{0.352}     \\ 
\bottomrule
\end{tabular}
\caption{The correlation between engagement scores and ground truth human judgment. Best scores are \textbf{emboldened} and second-best are \underline{underlined}. We train \textsc{EnDex} and \textsc{EnDex+NS} 10 times and report the mean with * and ** indicating a \texttt{stdev} < 0.05 and < 0.03, respectively. \textsc{EnDex-Best} is the best score observed over the 10 runs. Compared to existing metrics, the \textsc{EnDex}-framework achieves SOTA correlation with human judgement on engagingness, leading by far on our newly proposed \textsc{Red-Test} dataset with more complex and longer texts than chitchats.} 
\label{tab:main_results}
\end{table*}

\subsection{Experiment Set-up}
\label{subsec:setup}

We test the performance of the \textsc{EnDex} metric on 5 golden evaluation sets that have turn-level labels. Among them, \textsc{Better}~\citep{Ghazarian2019BetterAE}, \textsc{PredEeng-600}~\citep{Ghazarian2020PredictiveEA} are annotated specifically for engagingness with high annotator agreement. \textsc{Better} samples are taken from human conversation, while half of \textsc{PredEng-600} are chatbot generations. \textsc{FED}~\citep{Mehri2020UnsupervisedEO} annotates dialogue for 9 different dimensions, and we use their engagingness scores as target labels. \textsc{GRADE}~\citep{Huang2020GRADEAG} contains quality annotations for dialogue coherence, and we include it to test whether our model can also have good zero-shot performance on related tasks. Lastly, our own \textsc{RED-Test} is sourced from Reddit, contains discussions on various topics. A table of evaluation set statistics is provided in the Appendix\ref{tab:dataset_statistics}.

\subsection{Ablation Study on Engaging Dimensions} 
\label{subsec:ablation}
To test the robustness of the 4 engagingness dimensions \ref{sec:dims} of \textsc{EnDex}, we conducted ablation study to train model using only signals from each of the 4 dimensions. We hypothesize that dimensions with high positive contribution towards final results should have very successful clustering of engaging and non-engaging samples by itself; so, if we train model on data clustered by such dimension, we can still get good performance models. 





\begin{figure}[h]
    \centering
    \includegraphics[scale=0.17]{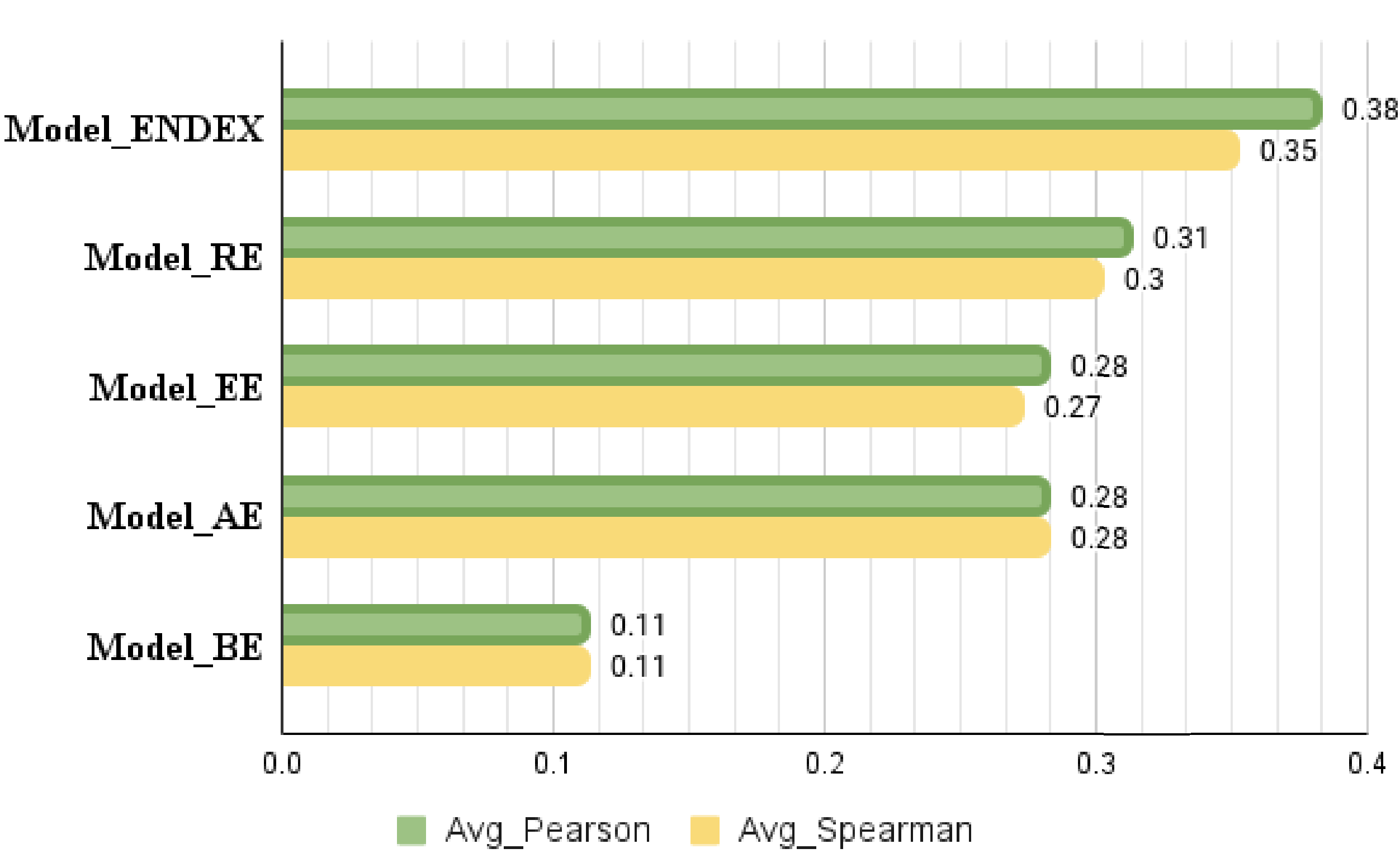}
    \caption{Ablation study of our four engagement dimensions. The \textsc{EnDex} model was trained on our aggregate engagingness score, while RE, EE, AE, and BE indicate models trains only on scores reflecting that particular dimension.}
    \label{fig:ablation_dimensions}
\end{figure}

We train five different models on different subsets of \textsc{RED}. All datasets included the same 40k negative (i.e. non-engaging) samples drawn according to our overall engagement score. However, the other 40k positive (i.e. engaging) samples were selected according to a particular dimension score (e.g. EE, AE, BE, and RE), except for \textsc{EnDex}, which is our aggregate score model. Figure \ref{fig:ablation_dimensions} shows that all four dimensions correlate with engagingness to some degree, but RE, AE, and EE are especially effective. We also observe a synergistic effect of training on a composite score rather than any one dimension individually. The experiment highlights and corroborates the multi-dimensionality of engagingness previously reported in the literature~\citep{See2019WhatMA}. Overall, having an aggregate score is crucial for successful distant-supervised annotation of negative and positive examples. 

\subsection{Comparison with Related Works}
\label{subsec:related_work_compare}

We compare our \textsc{EnDex} metric, and heuristics-augmented \textsc{EnDex+NS} metric with five baselines. Three baselines are rule-based, including \textbf{Random}, information \textbf{Specificity} ~\citep{See2019WhatMA} that counts number of non-stopword tokens, and \textbf{Inquisitiveness} ~\citep{Ghandeharioun2019ApproximatingIH} that examines question asking ability. We included them because in some dataset, rule-based system could work surprisingly well~\citep{Yeh2021ACA}.

We selected \textbf{USL-H} ~\citep{Phy2020DeconstructTR} as a baseline because it is the top performing metric on the \textsc{PredEng-600} and \textsc{Fed} engagingness evaluation sets~\citet{Yeh2021ACA}. USL-H is designed to measure high-level dialog quality, including understandability, sensibleness, and likability; it trains 3  BERT-based~\citep{bert} classifiers for each component, and uses a composite score named USL-H for overall assessment. \textsc{Pred\_En} ~\citep{Ghazarian2020PredictiveEA} uses BERT embedding plus MLP layer and train on ConvAI dataset~\citep{Dinan2019TheSC} to make engagement score predictions. \textbf{PRED\_EN (FT+DD)} further finetunes the original PRED\_EN metric on the DailyDialogue dataset, to get better results.

Our model has two versions: \textsc{EnDex} is solely trained on \textit{human-reaction} based data. \textsc{+NS} means non-engaging samples set is mixed with some rule-based negative samples, created by random insertion, random deletion, copying, and generic replies; 

The experiments in Table~\ref{tab:main_results} demonstrate that our model achieves strong performance on 4 engagingness related datasets, and good correlation with one coherence dataset(\textsc{Grade}). \textsc{EnDex} surpasses \textsc{Pred\_En} and USL-H by a large margin on two real human conversations, \textsc{Better} and \textsc{Red-Test}. USL-H still leads in \textsc{PredEng-600}, and \textsc{EnDex+NS}'s best model is a close second. \citet{Yeh2021ACA} shows achieving high score on \textsc{FED-eng} is challenging, with no one surpassing 0.3 spearman in 12 tested metrics. A strong rule-based question detection algorithm surprisingly claims the highest result, and \textsc{EnDex} a close second. 


We find that training solely with \textit{human reaction} distant supervision signals suffices for building competitive models on par or even surpassing mainstream metrics, and it shows better generalization capability in new domains, which seems to echo recent success on modeling human preferences via upvotes in Reddits~\citep{Gao2020DialogueRR}.







\section{Conclusion}
This paper proposes the first \textit{human reaction} based model, \textsc{EnDex}, to evaluate dialogue engagingness, and curates an 80k Reddit Engagement Dataset (\textsc{Red}) using a novel distant-supervision framework.
The success of \textsc{EnDex} demonstrates the validity of training automatic metrics with \textit{human reaction} signals, offering a strong complement to a synthetic negative sampling approach. We also release an off-the-shelf \textsc{EnDex} model, and a large scale dataset to facilitate future research.





\section*{Limitation}

One limitation is that we only curated data for turn-level dialogue. Multi-turn dialogues could also be useful, but it was computationally infeasible to interactively query Reddit for entire threads of conversation. Future work can explore this direction to produce dialogue-level and system-level engagingness metrics.

We also haven't fully explore our model's performance on non-dialogue domains, such as on story or creative generations. The training data distribution from the Reddit corpus is diverse enough that it could potentially achieve good performance in non-dialogue settings. A valuable direction of future work is to adapt our method for more general engagingness, or another evaluation metric for open-domain generation.

\section*{Ethics}
A caveat of using framing our approach around \textit{human attention} is that not all texts attracting high attention are good and ethical. Since being engaging often carries a positive connotation, we made a deliberate design decision to mitigate forms of \emph{negative engagement} in our metric. For example, we assign lower scores to samples flagged by Reddit as controversial, and our behavioral engagement dimension subtracts downvotes from upvotes to punish negative, biased~\citep{Liu_Jia_Wei_Xu_Wang_Vosoughi_2021}, and aggressive comments. Moreover, we implemented our emotional engagement algorithm to reward posts with positive emotional replies and punish posts that prompt negative emotions. Future may try to account for the darker aspects of engagingness into our framework and improve the \textsc{EnDex} metric to differentiate between positive and negative engagement. 


Human annotations for \textsc{RED-Test} were obtained via Amazon Mechanical Turks. We filtered out toxic samples to reduce the likelihood of offensive content and paid \$0.30 USD per instance for an expected hourly wage of \$20 USD.


\section*{Acknowledgement}
We want to give special thanks to Sarik Ghazarian for great advice and help with the baseline model; we also appreciate the brainstorming session with Hongyuan Yang and the suggestion to use z-score for sampling from Zeyi Wang.

\bibliography{anthology,custom}

\begin{thebibliography}{33}
\expandafter\ifx\csname natexlab\endcsname\relax\def\natexlab#1{#1}\fi

\bibitem[{Attfield et~al.(2011)Attfield, Kazai, and
  Piwowarski}]{Attfield2011TowardsAS}
Simon Attfield, Gabriella Kazai, and Benjamin Piwowarski. 2011.
\newblock Towards a science of user engagement ( position paper ).

\bibitem[{Chen et~al.(2005)Chen, Kolko, Cuddihy, and
  Medina}]{Chen2005ModelingAM}
Mark Chen, Beth~E. Kolko, Elisabeth Cuddihy, and Eliana Medina. 2005.
\newblock Modeling and measuring engagement in computer games.
\newblock In \emph{DiGRA Conference}.

\bibitem[{Demszky et~al.(2020)Demszky, Movshovitz-Attias, Ko, Cowen, Nemade,
  and Ravi}]{demszky-etal-2020-goemotions}
Dorottya Demszky, Dana Movshovitz-Attias, Jeongwoo Ko, Alan Cowen, Gaurav
  Nemade, and Sujith Ravi. 2020.
\newblock \href {https://doi.org/10.18653/v1/2020.acl-main.372}
  {{G}o{E}motions: A dataset of fine-grained emotions}.
\newblock In \emph{Proceedings of the 58th Annual Meeting of the Association
  for Computational Linguistics}, pages 4040--4054, Online. Association for
  Computational Linguistics.

\bibitem[{Devlin et~al.(2019)Devlin, Chang, Lee, and Toutanova}]{bert}
Jacob Devlin, Ming-Wei Chang, Kenton Lee, and Kristina Toutanova. 2019.
\newblock Bert: Pre-training of deep bidirectional transformers for language
  understanding.
\newblock \emph{ArXiv}, abs/1810.04805.

\bibitem[{Dinan et~al.(2019)Dinan, Logacheva, Malykh, Miller, Shuster, Urbanek,
  Kiela, Szlam, Serban, Lowe, Prabhumoye, Black, Rudnicky, Williams, Pineau,
  Burtsev, and Weston}]{Dinan2019TheSC}
Emily Dinan, Varvara Logacheva, Valentin Malykh, Alexander~H. Miller, Kurt
  Shuster, Jack Urbanek, Douwe Kiela, Arthur~D. Szlam, Iulian Serban, Ryan
  Lowe, Shrimai Prabhumoye, Alan~W. Black, Alexander~I. Rudnicky, Jason
  Williams, Joelle Pineau, Mikhail~S. Burtsev, and Jason Weston. 2019.
\newblock \href {https://arxiv.org/abs/1902.00098} {The second conversational
  intelligence challenge (convai2)}.
\newblock \emph{ArXiv preprint}, abs/1902.00098.

\bibitem[{Gao et~al.(2020)Gao, Zhang, Galley, Brockett, and
  Dolan}]{Gao2020DialogueRR}
Xiang Gao, Yizhe Zhang, Michel Galley, Chris Brockett, and Bill Dolan. 2020.
\newblock Dialogue response ranking training with large-scale human feedback
  data.
\newblock \emph{ArXiv}, abs/2009.06978.

\bibitem[{Ghandeharioun et~al.(2019)Ghandeharioun, Shen, Jaques, Ferguson,
  Jones, Lapedriza, and Picard}]{Ghandeharioun2019ApproximatingIH}
Asma Ghandeharioun, Judy~Hanwen Shen, Natasha Jaques, Craig Ferguson, Noah
  Jones, {\`{A}}gata Lapedriza, and Rosalind~W. Picard. 2019.
\newblock \href
  {https://proceedings.neurips.cc/paper/2019/hash/fc9812127bf09c7bd29ad6723c683fb5-Abstract.html}
  {Approximating interactive human evaluation with self-play for open-domain
  dialog systems}.
\newblock In \emph{Advances in Neural Information Processing Systems 32: Annual
  Conference on Neural Information Processing Systems 2019, NeurIPS 2019,
  December 8-14, 2019, Vancouver, BC, Canada}, pages 13658--13669.

\bibitem[{Ghazarian et~al.(2019)Ghazarian, Wei, Galstyan, and
  Peng}]{Ghazarian2019BetterAE}
Sarik Ghazarian, Johnny Wei, Aram Galstyan, and Nanyun Peng. 2019.
\newblock \href {https://doi.org/10.18653/v1/W19-2310} {Better automatic
  evaluation of open-domain dialogue systems with contextualized embeddings}.
\newblock In \emph{Proceedings of the Workshop on Methods for Optimizing and
  Evaluating Neural Language Generation}, pages 82--89, Minneapolis, Minnesota.
  Association for Computational Linguistics.

\bibitem[{Ghazarian et~al.(2020)Ghazarian, Weischedel, Galstyan, and
  Peng}]{Ghazarian2020PredictiveEA}
Sarik Ghazarian, Ralph~M. Weischedel, Aram Galstyan, and Nanyun Peng. 2020.
\newblock \href {https://aaai.org/ojs/index.php/AAAI/article/view/6283}
  {Predictive engagement: An efficient metric for automatic evaluation of
  open-domain dialogue systems}.
\newblock In \emph{The Thirty-Fourth {AAAI} Conference on Artificial
  Intelligence, {AAAI} 2020, The Thirty-Second Innovative Applications of
  Artificial Intelligence Conference, {IAAI} 2020, The Tenth {AAAI} Symposium
  on Educational Advances in Artificial Intelligence, {EAAI} 2020, New York,
  NY, USA, February 7-12, 2020}, pages 7789--7796. {AAAI} Press.

\bibitem[{Ghazarian et~al.(2022)Ghazarian, Wen, Galstyan, and
  Peng}]{Ghazarian2022DEAMDC}
Sarik Ghazarian, Nuan Wen, A.~G. Galstyan, and Nanyun Peng. 2022.
\newblock \href {https://arxiv.org/abs/2203.09711} {Deam: Dialogue coherence
  evaluation using amr-based semantic manipulations}.
\newblock \emph{ArXiv preprint}, abs/2203.09711.

\bibitem[{Hanu and {Unitary team}(2020)}]{Detoxify}
Laura Hanu and {Unitary team}. 2020.
\newblock Detoxify.
\newblock Github. https://github.com/unitaryai/detoxify.

\bibitem[{Huang et~al.(2020)Huang, Ye, Qin, Lin, and Liang}]{Huang2020GRADEAG}
Lishan Huang, Zheng Ye, Jinghui Qin, Liang Lin, and Xiaodan Liang. 2020.
\newblock \href {https://doi.org/10.18653/v1/2020.emnlp-main.742} {{GRADE}:
  Automatic graph-enhanced coherence metric for evaluating open-domain dialogue
  systems}.
\newblock In \emph{Proceedings of the 2020 Conference on Empirical Methods in
  Natural Language Processing (EMNLP)}, pages 9230--9240, Online. Association
  for Computational Linguistics.

\bibitem[{Lan et~al.(2020)Lan, Mao, Wei, Gao, and Huang}]{Lan2020PONEAN}
Tian Lan, Xian-Ling Mao, Wei Wei, Xiaoyan Gao, and Heyan Huang. 2020.
\newblock \href {https://arxiv.org/abs/2004.02399} {Pone: A novel automatic
  evaluation metric for open-domain generative dialogue systems}.
\newblock \emph{ArXiv preprint}, abs/2004.02399.

\bibitem[{Li et~al.(2016)Li, Galley, Brockett, Gao, and
  Dolan}]{li2015diversity}
Jiwei Li, Michel Galley, Chris Brockett, Jianfeng Gao, and Bill Dolan. 2016.
\newblock \href {https://doi.org/10.18653/v1/N16-1014} {A diversity-promoting
  objective function for neural conversation models}.
\newblock In \emph{Proceedings of the 2016 Conference of the North {A}merican
  Chapter of the Association for Computational Linguistics: Human Language
  Technologies}, pages 110--119, San Diego, California. Association for
  Computational Linguistics.

\bibitem[{Li et~al.(2017)Li, Su, Shen, Li, Cao, and
  Niu}]{li-etal-2017-dailydialog}
Yanran Li, Hui Su, Xiaoyu Shen, Wenjie Li, Ziqiang Cao, and Shuzi Niu. 2017.
\newblock \href {https://aclanthology.org/I17-1099} {{D}aily{D}ialog: A
  manually labelled multi-turn dialogue dataset}.
\newblock In \emph{Proceedings of the Eighth International Joint Conference on
  Natural Language Processing (Volume 1: Long Papers)}, pages 986--995, Taipei,
  Taiwan. Asian Federation of Natural Language Processing.

\bibitem[{Liu et~al.(2021)Liu, Jia, Wei, Xu, Wang, and
  Vosoughi}]{Liu_Jia_Wei_Xu_Wang_Vosoughi_2021}
Ruibo Liu, Chenyan Jia, Jason Wei, Guangxuan Xu, Lili Wang, and Soroush
  Vosoughi. 2021.
\newblock \href {https://ojs.aaai.org/index.php/AAAI/article/view/17744}
  {Mitigating political bias in language models through reinforced
  calibration}.
\newblock \emph{Proceedings of the AAAI Conference on Artificial Intelligence},
  35(17):14857--14866.

\bibitem[{Liu et~al.(2019)Liu, Ott, Goyal, Du, Joshi, Chen, Levy, Lewis,
  Zettlemoyer, and Stoyanov}]{Liu2019RoBERTaAR}
Yinhan Liu, Myle Ott, Naman Goyal, Jingfei Du, Mandar Joshi, Danqi Chen, Omer
  Levy, Mike Lewis, Luke Zettlemoyer, and Veselin Stoyanov. 2019.
\newblock \href {https://arxiv.org/abs/1907.11692} {Roberta: A robustly
  optimized bert pretraining approach}.
\newblock \emph{ArXiv preprint}, abs/1907.11692.

\bibitem[{Ma(2018)}]{inproceedings}
Xiaojuan Ma. 2018.
\newblock \href {https://doi.org/10.24963/ijcai.2018/809} {Towards
  human-engaged {AI}}.
\newblock In \emph{Proceedings of the Twenty-Seventh International Joint
  Conference on Artificial Intelligence, {IJCAI} 2018, July 13-19, 2018,
  Stockholm, Sweden}, pages 5682--5686. ijcai.org.

\bibitem[{Mcmahan(2003)}]{articlemcmahan}
Alison Mcmahan. 2003.
\newblock Immersion, engagement, and presence: A method for analyzing 3-d video
  games.
\newblock \emph{The Video Game Theory Reader}, pages 67--86.

\bibitem[{Mehri and Eskenazi(2020)}]{Mehri2020UnsupervisedEO}
Shikib Mehri and Maxine Eskenazi. 2020.
\newblock \href {https://aclanthology.org/2020.sigdial-1.28} {Unsupervised
  evaluation of interactive dialog with {D}ialo{GPT}}.
\newblock In \emph{Proceedings of the 21th Annual Meeting of the Special
  Interest Group on Discourse and Dialogue}, pages 225--235, 1st virtual
  meeting. Association for Computational Linguistics.

\bibitem[{Miller et~al.(2017)Miller, Feng, Batra, Bordes, Fisch, Lu, Parikh,
  and Weston}]{Miller2017ParlAIAD}
Alexander Miller, Will Feng, Dhruv Batra, Antoine Bordes, Adam Fisch, Jiasen
  Lu, Devi Parikh, and Jason Weston. 2017.
\newblock \href {https://doi.org/10.18653/v1/D17-2014} {{P}arl{AI}: A dialog
  research software platform}.
\newblock In \emph{Proceedings of the 2017 Conference on Empirical Methods in
  Natural Language Processing: System Demonstrations}, pages 79--84,
  Copenhagen, Denmark. Association for Computational Linguistics.

\bibitem[{Phy et~al.(2020)Phy, Zhao, and Aizawa}]{Phy2020DeconstructTR}
Vitou Phy, Yang Zhao, and Akiko Aizawa. 2020.
\newblock \href {https://doi.org/10.18653/v1/2020.coling-main.368} {Deconstruct
  to reconstruct a configurable evaluation metric for open-domain dialogue
  systems}.
\newblock In \emph{Proceedings of the 28th International Conference on
  Computational Linguistics}, pages 4164--4178, Barcelona, Spain (Online).
  International Committee on Computational Linguistics.

\bibitem[{Roller et~al.(2021)Roller, Dinan, Goyal, Ju, Williamson, Liu, Xu,
  Ott, Smith, Boureau, and Weston}]{Roller2021RecipesFB}
Stephen Roller, Emily Dinan, Naman Goyal, Da~Ju, Mary Williamson, Yinhan Liu,
  Jing Xu, Myle Ott, Eric~Michael Smith, Y-Lan Boureau, and Jason Weston. 2021.
\newblock \href {https://aclanthology.org/2021.eacl-main.24} {Recipes for
  building an open-domain chatbot}.
\newblock In \emph{Proceedings of the 16th Conference of the European Chapter
  of the Association for Computational Linguistics: Main Volume}, pages
  300--325, Online. Association for Computational Linguistics.

\bibitem[{Sato et~al.(2020)Sato, Akama, Ouchi, Suzuki, and
  Inui}]{Sato2020EvaluatingDG}
Shiki Sato, Reina Akama, Hiroki Ouchi, Jun Suzuki, and Kentaro Inui. 2020.
\newblock \href {https://doi.org/10.18653/v1/2020.acl-main.55} {Evaluating
  dialogue generation systems via response selection}.
\newblock In \emph{Proceedings of the 58th Annual Meeting of the Association
  for Computational Linguistics}, pages 593--599, Online. Association for
  Computational Linguistics.

\bibitem[{Savin-Baden et~al.(2014)Savin-Baden, Tombs, Bhakta, and
  Burden}]{SavinBaden2014StudentsEO}
Maggi Savin-Baden, Gemma Tombs, Roy Bhakta, and David J.~H. Burden. 2014.
\newblock Students’ experiences of emotional connection with pedagogical
  agents.

\bibitem[{Schoenau-Fog(2011)}]{SchoenauFog2011ThePE}
Henrik Schoenau-Fog. 2011.
\newblock The player engagement process - an exploration of continuation desire
  in digital games.
\newblock In \emph{DiGRA Conference}.

\bibitem[{See et~al.(2019)See, Roller, Kiela, and Weston}]{See2019WhatMA}
Abigail See, Stephen Roller, Douwe Kiela, and Jason Weston. 2019.
\newblock \href {https://doi.org/10.18653/v1/N19-1170} {What makes a good
  conversation? how controllable attributes affect human judgments}.
\newblock In \emph{Proceedings of the 2019 Conference of the North {A}merican
  Chapter of the Association for Computational Linguistics: Human Language
  Technologies, Volume 1 (Long and Short Papers)}, pages 1702--1723,
  Minneapolis, Minnesota. Association for Computational Linguistics.

\bibitem[{Sellam et~al.(2020)Sellam, Das, and Parikh}]{Sellam2020BLEURTLR}
Thibault Sellam, Dipanjan Das, and Ankur Parikh. 2020.
\newblock \href {https://doi.org/10.18653/v1/2020.acl-main.704} {{BLEURT}:
  Learning robust metrics for text generation}.
\newblock In \emph{Proceedings of the 58th Annual Meeting of the Association
  for Computational Linguistics}, pages 7881--7892, Online. Association for
  Computational Linguistics.

\bibitem[{Silpasuwanchai et~al.(2016)Silpasuwanchai, Ma, Shigemasu, and
  Ren}]{inproceedings_Silpasuwanchai}
Chaklam Silpasuwanchai, Xiaojuan Ma, Hiroaki Shigemasu, and Xiangshi Ren. 2016.
\newblock \href {https://doi.org/10.1145/2901790.2901836} {Developing a
  comprehensive engagement framework of gamification for reflective learning}.
\newblock pages 459--472.

\bibitem[{Tao et~al.(2018)Tao, Mou, Zhao, and Yan}]{Tao2018RUBERAU}
Chongyang Tao, Lili Mou, Dongyan Zhao, and Rui Yan. 2018.
\newblock \href
  {https://www.aaai.org/ocs/index.php/AAAI/AAAI18/paper/view/16179} {{RUBER:}
  an unsupervised method for automatic evaluation of open-domain dialog
  systems}.
\newblock In \emph{Proceedings of the Thirty-Second {AAAI} Conference on
  Artificial Intelligence, (AAAI-18), the 30th innovative Applications of
  Artificial Intelligence (IAAI-18), and the 8th {AAAI} Symposium on
  Educational Advances in Artificial Intelligence (EAAI-18), New Orleans,
  Louisiana, USA, February 2-7, 2018}, pages 722--729. {AAAI} Press.

\bibitem[{Yeh et~al.(2021)Yeh, Esk{\'e}nazi, and Mehri}]{Yeh2021ACA}
Yi-Ting Yeh, Maxine Esk{\'e}nazi, and Shikib Mehri. 2021.
\newblock \href {https://arxiv.org/abs/2106.03706} {A comprehensive assessment
  of dialog evaluation metrics}.
\newblock \emph{ArXiv preprint}, abs/2106.03706.

\bibitem[{Zhang et~al.(2018)Zhang, Dinan, Urbanek, Szlam, Kiela, and
  Weston}]{Zhang2018PersonalizingDA}
Saizheng Zhang, Emily Dinan, Jack Urbanek, Arthur Szlam, Douwe Kiela, and Jason
  Weston. 2018.
\newblock \href {https://doi.org/10.18653/v1/P18-1205} {Personalizing dialogue
  agents: {I} have a dog, do you have pets too?}
\newblock In \emph{Proceedings of the 56th Annual Meeting of the Association
  for Computational Linguistics (Volume 1: Long Papers)}, pages 2204--2213,
  Melbourne, Australia. Association for Computational Linguistics.

\bibitem[{Zhang et~al.(2020)Zhang, Kishore, Wu, Weinberger, and
  Artzi}]{zhang2019bertscore}
Tianyi Zhang, Varsha Kishore, Felix Wu, Kilian~Q. Weinberger, and Yoav Artzi.
  2020.
\newblock \href {https://openreview.net/forum?id=SkeHuCVFDr} {Bertscore:
  Evaluating text generation with {BERT}}.
\newblock In \emph{8th International Conference on Learning Representations,
  {ICLR} 2020, Addis Ababa, Ethiopia, April 26-30, 2020}. OpenReview.net.

\end{thebibliography}
\bibliographystyle{acl_natbib}

\appendix

\section{Appendix}
\label{sec:appendix}





\subsection{\textsc{RED} Processing Steps}
Our Reddit data is downloaded from Pushshift.io \footnote{https://files.pushshift.io/reddit/comments/}, and we processed approximately 5M data to curate an 80k sample \textsc{RED} dataset. We deleted posts that do not have an immediate parent thread, because we need pair turn-level data. Our preprocessing removes non-conversational data, such as posts including \& gt(reply to symbol). We also removed explicitly toxic data filtered by Detoxify~\citep{Detoxify}.  

We also applied a key data processing trick to reduce noisy signals -- the exposure variable. It helps measure the amount exposure each post receives to help normalize its upvote/reply score. We reward posts that are in low-exposure, unpopular threads, while penalizing posts in high-exposure, popular threads, because high upvotes and replies in popular threads may be more due to exposure than true engagingness. 

After computing the normalized score given in Equation \ref{eqn:eqn}, we also apply another z-score to normalize the final \textsc{EnDex} score according to standard deviation, so that we can easily sample our data from it. A score with higher standard deviation will imply a higher probability that the sample is engaging according to our metric. We apply a cut-off to sample high probability engaging and non-engaging samples, and arrive at the \textsc{RED} dataset.



\subsection{Model training details}

Our \textsc{RED-Test} set contains 300 human labeled data. The train validation split during training is 0.8 and 0.2. 

We used 4 Nvidia A6000 GPUs for training, and 1 Nvidia A6000 GPU for inference. The average runtime for training one model is 2 minutes per epoch, and inference time is in seconds, negligible for the testset. The estimated energy cost per model is, assuming per second gpu energy cost of 245W: 245W*4*60 = 58800 per model.

We trained our model for 2 epochs, and only save the best checkpoints, with learning rate of 5e-5 with no extensive hyperparameter search. 

We used specificity and question examination inspired from \citep{See2019WhatMA}; USL-H \citep{Phy2020DeconstructTR} and \textsc{PredEn} is taken from a GitHub repo\footnote{https://github.com/exe1023/DialEvalMetrics} and modified to use a local \texttt{bert-base-uncased} since the original `bert-as-service` code no longer functions.




The Formula for calculating each dimension is given in the following:

 \begin{itemize}

 


   \item \textbf{Reply Engagement}; 
The raw Reply Engagement score(re) is just the reply counts of a post. Popularity value adjusted RE score is given by PVRE in equation (\ref{eq:pvre}), where $M_{pv}$ and $M_{re}$ are the median of popularity value and reply counts in the entire dataset. Please refer to equation (\ref{eq:pv}) for calculation of the popularity value.

 \begin{equation}
 \small
    \label{eq:pvre}
     \textrm{PVRE}(re) = re + \frac{M_{pv}}{M_{\textrm{re}}}  *  \frac{re}{\textrm{PV(re)}} * re
 \end{equation}

 \item \textbf{Behavioral Engagement}; 
The raw BE score of a certain post(be) is obtained by subtracting downvotes to upvotes and set to 0 if given controversy flag. Popularity value adjusted BE score is given by PVBE in equation (\ref{eq:pvbe}), where $M_{pv}$ and $M_{be}$ are the median of popularity value and raw BE score in the entire dataset.  Please refer to equation (\ref{eq:pv}) for calculation of the popularity value.

 \begin{equation}
 \small
    \label{eq:pvre}
     \textrm{PVBE}(be) = be + \frac{M_{pv}}{M_{\textrm{be}}}  *  \frac{be}{\textrm{PV(be)}} * be
 \end{equation}

 \item \textbf{Attentional Engagement}; 
It is calculated using maximum information specificity, or the the maximum number of non-stopword tokens in a post's replies, and whether its children posts are editted; t is the maximum reply specificity, and e stands for number of edited replies. 

 \begin{equation}
     \textrm{AE}(x) = t + 10 * e
 \end{equation}



 \item \textbf{Emotional Engagement}; 
The EE score is the aggregate probability for all positive emotion categories, produced by the go-emotion classifier~\citep{demszky-etal-2020-goemotions}. 

 \end{itemize}

\subsection{Submodular Normalization and z-score Clustering}
After we obtained the sub-dimension scores, we want to aggregate them into a single normalized \textsc{EnDex} Score, and lastly cluster them into positive and negative sets to train a binary classifier. The formulas are list in the following: \\

\begin{equation}
\small
    \label{eqn:eqn}
     \textsc{EnDex}(x) = \Bigg(\displaystyle\sum_{n=1} ^{3} w_i * \frac{x_i}{x_i+ \alpha_i} \Bigg) +  w_{\textrm{EE}} * x_{\textrm{EE}},
\end{equation}

\noindent $N$ is the normalized score for sample $x$, $x_i$ is $x$'s raw score on dimension $i$, where $i\in \{RE,BE,AE\}$; $\alpha_i$ is the median of $i$-th dimension; $w_i$ is the weight for $i$ dimension; $w_{\textrm{EE}}$ is the weight for EE dimension. The weight can be tuned for your own usage of \textsc{RED};\\

A confidence threshold $\kappa$(ours is 1) needs to be picked, which means that we regard samples that fall between $\kappa$ standard deviation from mean as uncertain, and are thus discarded. And we cluster positive and negative samples using the following equation (\ref{eq:zscore}).\\

\begin{equation}
\label{eq:zscore}
\small Polarity(x) = \begin{cases}
1 & \text{if } z\_score(x)>\kappa \\
0 & \text{if } z\_score(x)<-\kappa 
\end{cases}
\end{equation}

\begin{figure*}[!h]
    \centering
    \includegraphics[width=0.98\textwidth]{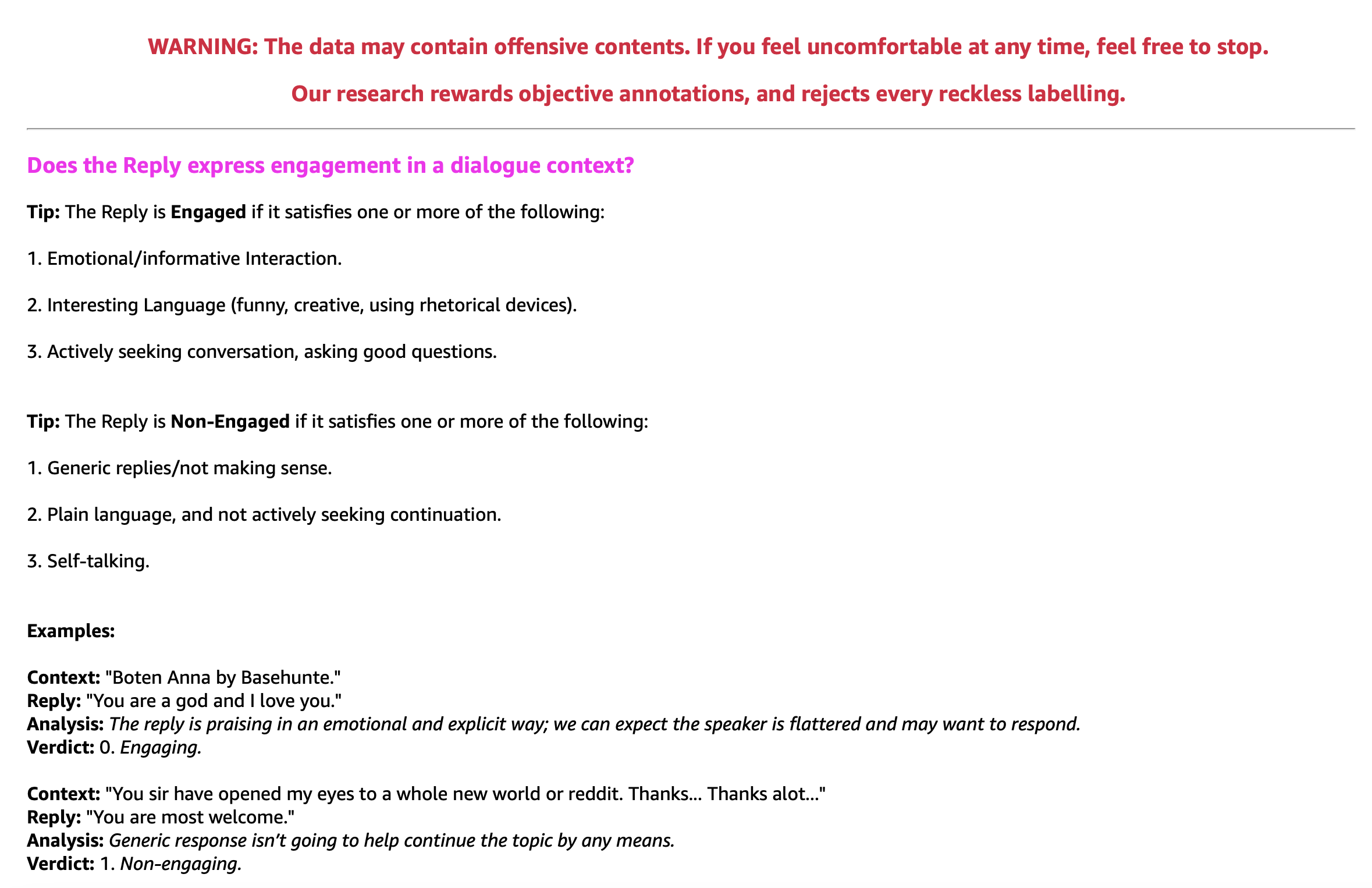}
    \caption{The screenshot of the task description of our Amazon MTurk questionnaire. We have prepared instructions, demonstrations, and proper warning of offensive content.}
    \label{fig:mturk}
\end{figure*}

\begin{figure*}[!h]
    \centering
    \includegraphics[width=0.98\textwidth]{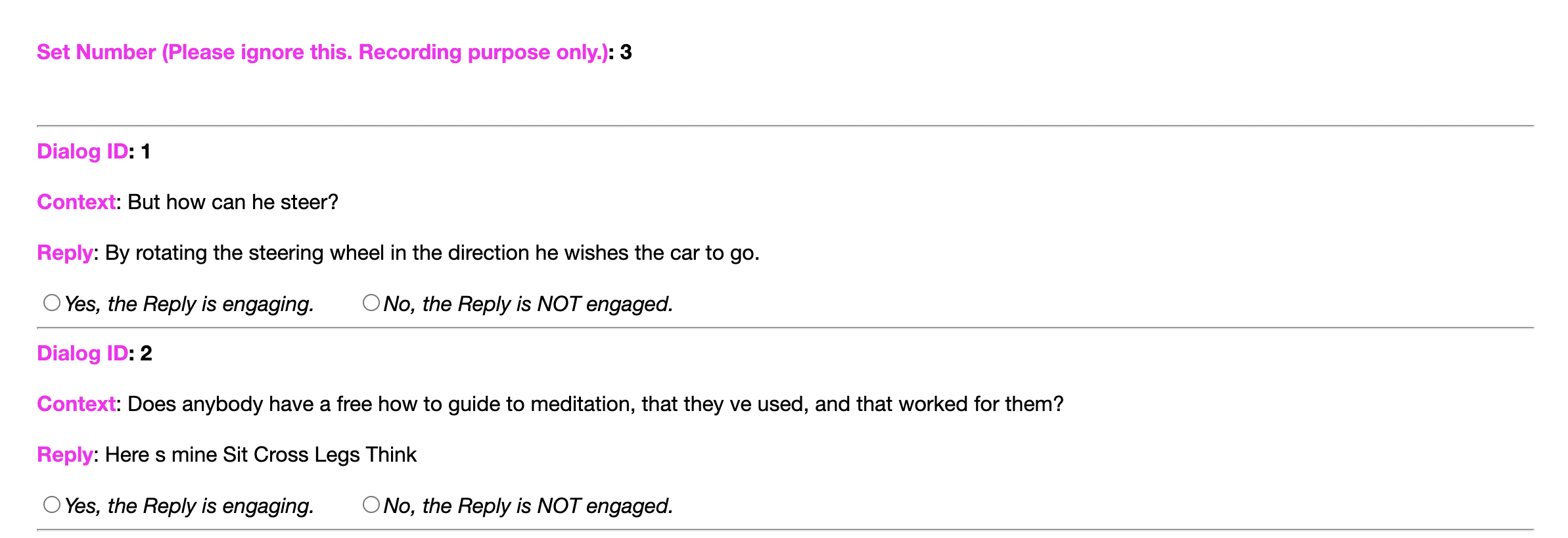}
    \caption{The screenshot of the labeling area of our Amazon MTurk questionnaire. Each pair will be labelled by three annotators.}
    \label{fig:label}
\end{figure*}

\subsection{Annotation data and test data}
We performed annotation on Amazon Mechanical Turk, and selected annotators based in the United State; in implemented restrictions to annotator with 98\% pass rate. We give four examples and clear instruction for the task carried out. A screenshot of our annotation interface is provided below.

Table \ref{tab:dataset_statistics} gives summary of the evaluation datasets we used.


 
 
 
 
 

\begin{table*}[!hb]
\centering
\resizebox{\textwidth}{!}{%
\begin{tabular}{lccclc}
\toprule
\textbf{Dataset} & \textbf{\# of Samples} & \textbf{Context Length} & \textbf{Response Length} & \multicolumn{1}{c}{\textbf{Source}} & \textbf{Agreement Rate} \\
\midrule
BETTER           & 297 & 6  & 8  &  Human   & N/A  \\
PREDENG-600      & 600 & 12 & 14 & Human+Bot  & 0.51 \\
FED-ENG          & 261 & 26 & 12 & Human+Bot  & N/A \\
RED-TEST         & 150 & 16   & 17   &  Human    & 0.34 \\
GRADE            & 150 & 12 & 14 & Human  & N/A  \\
\bottomrule
\end{tabular}%
}
\caption{Dataset Statistics for the 5 golden evaluation sets, with number of samples, context-length, response length, and  if applicable, inter-annotator agreement rate.}
\label{tab:dataset_statistics}
\end{table*}

\end{document}